# An Acceleration Scheme to The Local Directional Pattern

Y.M. Ayami
Durban University of Technology
Department of Information
Technology, Ritson Campus, Durban,
South Africa
ayamlearning@gmail.com

A. Shabat
Durban University of Technology
Department of Information
Technology, Ritson Campus,
Durban,
South Africa
abshabat@gmail.com

## ABSTRACT

This study seeks to improve the running time of the Local Directional Pattern (LDP) during feature extraction using a newly proposed acceleration scheme to LDP. LDP is considered to be computationally expensive. To confirm this, Shabat and Tapamo compared the running time of the LDP to gray level co-occurrence matrix (GLCM) were it was established that the running time for LDP was two orders of magnitude higher than that of the GLCM. In this study, the performance of the newly proposed acceleration scheme was evaluated against LDP and LBP using images from the publicly available extended Cohn-Kanade (CK+) dataset. Based on our findings, the proposed acceleration scheme significantly improves the running time of the LDP by almost 3 times during feature extraction.

## 1. INTRODUCTION

Texture analysis has been an active research area since 1960. This has resulted to the development and existence of a wide range of techniques over time which have been proposed to discriminate textures. Prior to the existence of local texture descriptors, Pietikäinen *et al*. [2] asserted that most of the proposed methods were not capable of meeting the requirements of many applications due to their poor discriminative performance on real-world textures and computation complexity. To overcome these challenges, local texture descriptors were introduced [3]. Examples of local descriptors include Directional Local Binary Pattern [4], Enhanced Local Directional Patterns (ELDP) [5], Local Binary Pattern (LBP) [6] and Local Directional Pattern (LDP) [7]. These descriptors have been deployed in a variety of applications amongst others palmprint recognition, face recognition [8] and facial expression analysis [9].

The LDP is a feature extraction method introduced by Jabid, Kabir and Chae [7] to overcome the challenges of the Local Binary Pattern (LBP) such as noise and illumination change. Despite being robust in the presence of noise, the LDP is considered to be computationally expensive. Shabat and Tapamo [1] compared the running time of the LDP to gray level co-occurrence matrix (GLCM) were it was established that the running time for LDP was two orders of magnitude higher than that of the GLCM. The authors further asserted that this is because certain calculations on the kirsch mask which produce same results are repeated when performing calculations. Meanwhile, the LBP is said to be computationally effiecient.

Against this background, this study proposes a novel approach which seeks to accelerate the running time of the LDP. This acceleration scheme omits redundant calculations consequently, improving the running time of the LDP. To the best of our knowledge, no studies have been conducted thus far which aim at improving the speed of the LDP. The major contribution of this study is an acceleration scheme to the LDP which reduces its running time by almost 3 times.

The rest of the study is categorized as follows: Section Two presents the Local features for texture analysis which includes the LBP, the LDP and the proposed acceleration scheme of the LDP calculation. Section Three explains how the experiments and Section Four presents the experimental findings and discussion. Section Five concludes this study.

## 2. LOCAL FEATURES FOR TEXTURE ANALYSIS

### 2.1 Local Binary Pattern (LBP)

In 1996, Ojala et al, proposed the LBP, a local texture descriptor which is considered to have a low computational complexity and high discriminative power [10]. The LBP works by assigning a label to every pixel of an image by thresholding the $3 \times 3$ neighborhood of each pixel with the center pixel value and considering the result as a binary number. A 256-bin histogram of LBP labels computed over the region is used as a texture descriptor.

### 2.2 Local Directional Pattern (LDP)

The LDP is based on the known Kirsch kernels. It is based on eight different directions wherein the edge response values are considered [11]. The LDP features are composed of an eight-bit binary code. Each pixel of an input image is assigned to this code. The following are the three steps which are used to generate this code:

### 2.2.1 Calculation of eight directional responses.

The Figure 1 depicts the calculation of the eight directional responses of a particular pixel of an image. The Kirsch compass edge detector in eight orientations ($M_0, M_1, ..., M_7$) centered on its own position is used to facilitate this calculation.

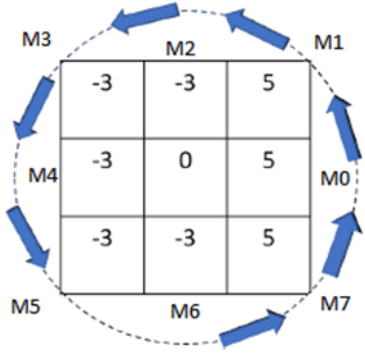

Figure 1: The Kirsch masks in eight different directions

The directional response $M_i$ for every pixel (x,y) of an input image, $I$, is computed using the equation 1 where is the direction and $M_i$ represents the corresponding mask.

$$M_i = \sum_{k=-1}^{1} \sum_{l=-1}^{1} M_1(k+1, l+1) \times I(x+k, y+l) \quad (1)$$

After computing the directional responses in all the eight directions, the ($m_7, m_6, ..., m_0$) is derived.

### 2.2.2. LDP code generation.

Based on the directional derived obtained in the previous step (step 2.2.1). The three most outstanding responses k are selected and their corresponding bit are set to 1 leaving the rest (8 - k) bits set to 0. Lastly, the LDP code for image I is generated, $LDP_{x,y}(m_0, m_1, ..., m_7)$, of the pixel (x,y) with the eight directional response ($m_0, m_1, ..., m_7$) is generated using Equation 2.

$$LDP_{x,y}(m_0, m_1, ..., m_7) = \sum_{i=0}^{7} s(m_i - m_k) \times 2^i \quad (2)$$

where $m_k$ represents the $k^{th}$ most outstanding responses and $s(x)$ is defined as:

$$s(x) = \begin{cases} 1, & if\ x \geq 0 \\ 0, & otherwise \end{cases} \quad (3)$$

### 2.2.3. Construction of LDP descriptor.

Finally, after the calculation of the LDP code for each pixel ($x, y$), the LDP descriptor is constructed. The LDP histogram is then used to represent the input image $I$ of size $m \times n$ using equation 4. More often, the value of $k$ is 3 ($k = 3$); this means that $8_{C_3}$ = 56 distinct values are generated which are then used to encode an image. These 56 distinct values are represented by a histogram $H$ which be defined as:

$$H_i = \sum_{x=0}^{M-1} \sum_{y=0}^{N-1} p(LDP_{(x,y)}, C_i) \quad (4)$$

where $C_i$ is represents the $i^{th}$ LDP pattern value, $i = 1, ..., 8_{C_3}$ and $p$ is defined in the Equation 5.

$$p(x, a) = f(x) = \begin{cases} 1, & if\ x = a \\ 0, & if\ x \neq 0 \end{cases} \quad (5)$$

For every texture, $T$, an LDP feature vector $LDP_{P,T}$ is extracted and denoted as equation 6 where $k$ represents number of bits which are most significant.

$$ldp_{k,T} = (H_1, H_2, ..., H_{56}) \quad (6)$$

## 2.3 Acceleration of the LDP Calculation (ALDP)

This section proposes a scheme that significantly reduces the running time during the computation convolution of an image with the Kirsch masks. Currently, the LDP works by multiplying a $3 \times 3$ matrix with the eight Kirsch edge response masks. However, certain rows and columns in the kirsch are redundant. As can be noticed in Figure 2, the first columns of $M_0$, $M_1$ and $M_7$ are the same. The drawback to this is that, whenever calculation operations such as multiplication are performed on these redundant columns, the same results are obtained. This in return adds onto the computation time on each operation and thus, making the computation convolution of an image even slower. The Figure 2 shows the kirsch mask with the redundant columns highlighted.

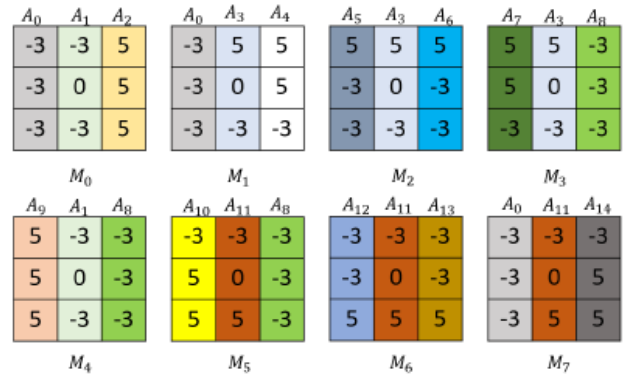

Figure 2: The eight Kirsch masks with the redundant columns highlighted. Examples of redundant columns include the first columns in $M_o$, $M_1$ and $M_7$

Given an image $I = (I_{ij})$ where $0 \leq i \leq R - 1$ and $0 \leq j \leq C - 1$, $R$ is the number of rows and $C$ the number of columns. For each pixel $(x, y)$ of the image $I$, the computation of directional responses $(m_0, m_1, ..., m_7)$, from the convolution with the Kirsch masks is decomposed into 2 steps as shown in equations 7 to 29. These two steps include the following:

### 2.3.1. Kirsch Mask Reconstruction.

During this step, each unique column of the kirsch mask is decomposed into equations 7 to 21. Thereby omitting the redundant columns. This step provides the following unique equations which are a column decomposition of the kirsch mask.

$$A0 = -3(I_{(x-1,y-1)} + I_{(x,y-1)} + I_{(x+1,y-1)}) \quad (7)$$
$$A1 = -3(I_{(x-1,y)} + I_{(x+1,y)}) \quad (8)$$
$$A2 = 5(I_{(x-1,y+1)} + I_{(x,y+1)} + I_{(x+1,y+1)}) \quad (9)$$
$$A3 = 5(I_{(x-1,y)}) - 3(I_{(x+1,y)}) \quad (10)$$
$$A4 = 5(I_{(x-1,y+1)} + I_{(x,y+1)}) + (-3(I_{(x+1,y+1)})) \quad (11)$$
$$A5 = 5(I_{(x-1,y-1)}) + (-3(I_{(x,y-1)} + I_{(x+1,y-1)})) \quad (12)$$
$$A6 = 5(I_{(x-1,y+1)}) + (-3(I_{(x,y+1)} + I_{(x+1,y+1)})) \quad (13)$$
$$A7 = 5(I_{(x-1,y-1)}) + (-3(I_{(x+1,y-1)})) \quad (14)$$
$$A8 = -3(I_{(x-1,y+1)} + I_{(x,y+1)} + I_{(x+1,y+1)}) \quad (15)$$
$$A9 = 5(I_{(x-1,y-1)} + I_{(x,y-1)}) + (-3(I_{(x+1,y-1)})) \quad (16)$$
$$A10 = -3(I_{(x-1,y-1)}) + 5(I_{(x,y-1)} + I_{(x+1,y-1)}) \quad (17)$$
$$A11 = -3(I_{(x-1,y)} + I_{(x,y)}) + 5(I_{(x+1,y-1)}) \quad (18)$$
$$A12 = -3(I_{(x-1,y-1)} + I_{(x,y-1)}) + 5(I_{(x+1,y+1)}) \quad (19)$$
$$A13 = -3(I_{(x-1,y+1)} + I_{(x,y+1)}) + 5(I_{(x+1,y+1)}) \quad (20)$$
$$A14 = -3(I_{(x-1,y+1)}) + 5(I_{(x,y+1)} + I_{(x+1,y+1)}) \quad (21)$$

### 2.3.2. Edge Response Calculation.

In this step, the Kirsch edge responses $m_0, ..., m_7$ are calculated using the reconstructed equations 7 to 21 which were derived through the decomposition of the kirsch mask.

### 2.3.3. Algorithm Analysis.

For an image of size $\times m$, ALDP running time will be $A(m,n) = 0(mn)$. However, the running time for LDP is $L(m,n) = R(m,n) + H(m,n)$ where $R(m,n) = 0(mn)$ is the running time to compute the responses and $H(m,n) = 0(mn)$ is the running time of the computation of the histogram, then $L(m,n) = 0(mn)$.

Table 1 shows how the proposed accelerated scheme significantly reduced the number of multiplications during the computation of the kirsch edge responses.

$$m_0 = A_0 + A_1 + A_2 \quad (22)$$
$$m_1 = A_0 + A_3 + A_4 \quad (23)$$
$$m_2 = A_5 + A_3 + A_6 \quad (24)$$
$$m_3 = A_7 + A_3 + A_8 \quad (25)$$
$$m_4 = A_9 + A_1 + A_8 \quad (26)$$
$$m_5 = A_{10} + A_{11} + A_8 \quad (27)$$
$$m_6 = A_{12} + A_{11} + A_{13} \quad (28)$$
$$m_7 = A_0 + A_{11} + A_{14} \quad (29)$$

Table 1: The number of multiplications and additions for each extraction method

| Extraction Method | Multiplication | Addition |
|---|---|---|
| ALDP | 30 | 46 |
| LDP | 72 | 64 |
| LBP | 0 | 8 |

## 3. EXPERIMENTAL SETUP

### 3.1. Data Set

This study made use of images from the publicly available CK+ dataset of size $(640 \times 490)$ pixels. The viola-Jones detection technique was employed to detect the face and susbsequntly the images were resized to $(260 \times 260)$ pixels were used.

### 3.2. Sliding Window

This study also examined the effect of windowing during feature extraction. Sliding windows are applied to an image inorder to increase the number of features extracted from an image subsequently boasting the accuracy of the classifiers during classification. To achieve this, the following window sizes were used $((10 \times 10), (20 \times 20), (25 \times 25), (50 \times 50)$ and $(100 \times 100))$ pixels.

### 3.3. Computing Details

Experiments for this study were conducted on a computer with the following specifications: 6 gigabytes of RAM, Intel Core i5-3230M Processor and a CPU with speed 2.60GHz. These experiments were implemented using Python framework, Scikitlearn library and OpenCV.

## 4. FINDINGS AND DISCUSSION

### 4.1. Running time of Local Descriptors

In Table 2, the running times of local descriptors (LBP and LDP) were compared to the running time of the newly proposed acceleration scheme to the LDP. During the experiment, the running times were tested using different number of images. As the number of images increased, so did the running time. As can be observed in Figure 3, LDP took so long to extract features compared to LBP and the proposed acceleration scheme.

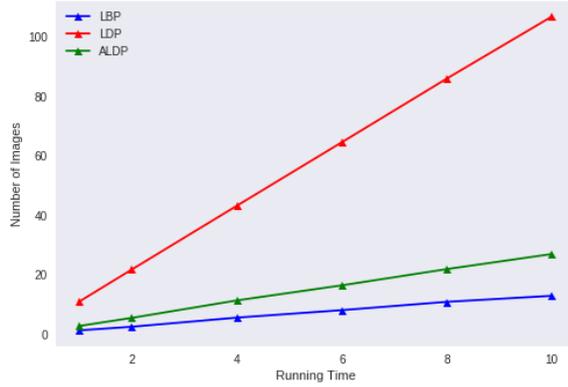

Figure 3: The running times of LBP, LDP and the ALDP are compared using various number of images.

Based on the findings in Table 2, in agreement to the assertions of Shabat and Tapamo [1], the LDP is computationally expensive. It took the LDP 11.43 seconds to extracts feature from a single image where as the the proposed acceleration scheme only took 3.61 seconds and only 1.87 seconds for the LBP. This clearly confirms that the proposed acceleration scheme outperforms the LDP by almost three times. It can therefore be deduced from the findings of this study that the proposed acceleration scheme significantly reduced the running time of the LDP by almost three times.

Table 2: The running times of LBP, LDP and the accelerated LDP (ALDP) are compared using various number of images.

| Number of Images | Local Descriptors | | |
|---|---|---|---|
| | LBP (s) | LDP (s) | ALDP (s) |
| 1 | 1.33 | 11.04 | 2.81 |
| 2 | 2.54 | 21.79 | 5.52 |
| 4 | 5.6 | 43.18 | 11.38 |
| 6 | 8.11 | 64.49 | 16.44 |
| 8 | 10.9 | 85.86 | 21.89 |
| 10 | 12.92 | 106.67 | 26.95 |

### 4.2. Effect of sliding window on running time.

This study focused on the acceleration of the LDP during feature extraction. However, the study also examined the effect of applying a sliding window to an image during the feature extraction process using different number of windows sizes. The Table 2 describes the results that were recorded during the experiment. The findings show that, having a small window size results into having a larger number of sliding windows. Furthermore, it was observed that, the larger the number of sliding windows, the lesser the running time during feature extraction. For instance, when the sliding window was $10 \times 10$ pixels, the proportionate number of sliding windows was $784$. With this number of sliding windows, LBP took $0.97$ seconds, LDP took $7.28$ seconds where as the newly proposed ALDP took 1.29 seconds. After adjusting the sliding window size to $25 \times 25$ pixels, the number of the resulting window sizes from the image reduced from 784 to 144. This adjustment resulted into the increase in the running time during feature extraction by the local descriptors. Subsquently, the LBP's running time increased to 1.13 seconds , 9.28 seconds for the LDP and 2.42 seconds for ALDP. It therefore can be deduced that the sliding window has a significant impact on the running time of the extraction methods.

Table 3: The running times of LBP, LDP and the accelerated LDP (ALDP) are compared with various window sizes applied.

| window size | number of sliding windows | Local Descriptors | | |
|---|---|---|---|---|
| | | LBP (s) | LDP (s) | ALDP (s) |
| $10 \times 10$ | 784 | 0.97 | 7.28 | 1.29 |
| $20 \times 20$ | 196 | 1.1 | 9.04 | 2.36 |
| $25 \times 25$ | 144 | 1.13 | 9.28 | 2.42 |
| $50 \times 50$ | 36 | 1.2 | 10.1 | 2.6 |
| $100 \times 100$ | 9 | 1.24 | 10.53 | 2.7 |

### 5. CONCLUSION

This study proposed ALDP, a novel acceleration scheme to the LDP, an improvement to its running time. To test this improvement, the LBP, LDP and the newly proposed accelerated LDP were tested on images from the CK+ dataset. Results show that the proposed acceleration scheme made a significant improvement to the running time of the LDP. The study also examined the effect of applying a sliding window to the image during feature extraction. Based on the finding, it was established that windowing has an impact on the running time during feature extraction. Future work could focus on identifying more patterns in this newly proposed acceleration scheme. These patterns could further reduce the running time and subsequently save on the usage of computing resources during feature extraction with the LDP.